\documentclass{article}
\usepackage{spconf,amsmath,graphicx,url,booktabs,multirow,caption,subcaption,bm,bbm}

\newcommand{\one}[1]{\mathbbm{1}_{[#1]}}

\title{Supervising Remote Sensing Change Detection Models with \\3D Surface Semantics}

\name{Isaac Corley and Paul Rad}
\address{University of Texas at San Antonio}

\name{Isaac Corley \qquad Peyman Najafirad}

  \address{Secure Artificial Intelligence Laboratory for Autonomy (AILA) \\
                            The University of Texas at San Antonio, Texas, USA \\
                          \{isaac.corley,  peyman.najafirad\}@utsa.edu\\
                          \thanks{$^*$ Corresponding author}}

\begin{document}

\maketitle

\begin{abstract}
Remote sensing change detection, identifying changes between scenes of the same location, is an active area of research with a broad range of applications. Recent advances in multimodal self-supervised pretraining have resulted in state-of-the-art methods which surpass vision models trained solely on optical imagery. In the remote sensing field, there is a wealth of overlapping 2D and 3D modalities which can be exploited to supervise representation learning in vision models. In this paper we propose Contrastive Surface-Image Pretraining (CSIP) for joint learning using optical RGB and above ground level (AGL) map pairs. We then evaluate these pretrained models on several building segmentation and change detection datasets to show that our method does, in fact, extract features relevant to downstream applications where natural and artificial surface information is relevant.\footnote{Experimental code is available at \url{https://github.com/isaaccorley/contrastive-surface-image-pretraining} and model checkpoints are made available in the TorchGeo \cite{stewart2021torchgeo} library at \url{https://github.com/microsoft/torchgeo}}

\end{abstract}


\section{Introduction}
\label{sec:intro}
\begin{figure}[ht!]
  \centering
  \centerline{\includegraphics[width=\linewidth]{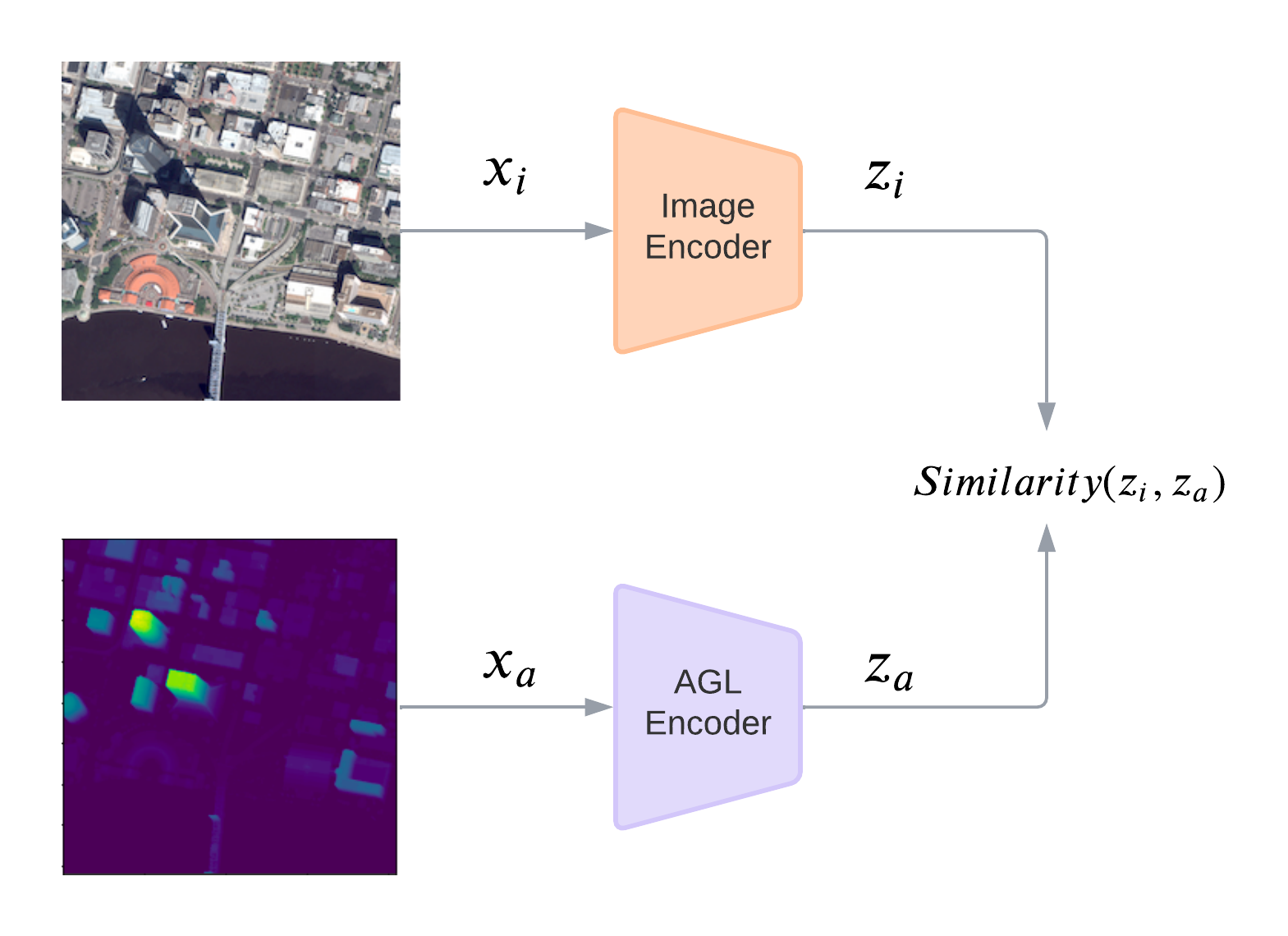}}
\caption{Our Contrastive Surface-Image Pre-training (CSIP) model learns the combined latent distribution, 3D surface semantics while performing similarity matching of optical RGB and AGL map pairings.}
\label{fig:csip}
\end{figure}

Self-supervised learning methods have recently become the de facto pretraining method due to their conceptual simplicity and ability to make use of large unlabeled datasets. Due to the wealth of multiview and multimodal data, self-supervision methods have shown dramatic improvements in the remote sensing and earth observation fields. Manas et al. \cite{manas2021seasonal} showed improved performance while pretraining using Sentinel-2 satellite imagery of the same location from different seasons as multiple views. Heidler et al. \cite{heidler2021self} experimented with pretraining on an audiovisual dataset of aerial image and audio pairs and showed efficient learning and improved downstream performance compared to transfer learning from ImageNet as well as other state-of-the-art self-supervised learning methods.

\begin{figure*}[t!]
  \centering
  \centerline{\includegraphics[width=\textwidth]{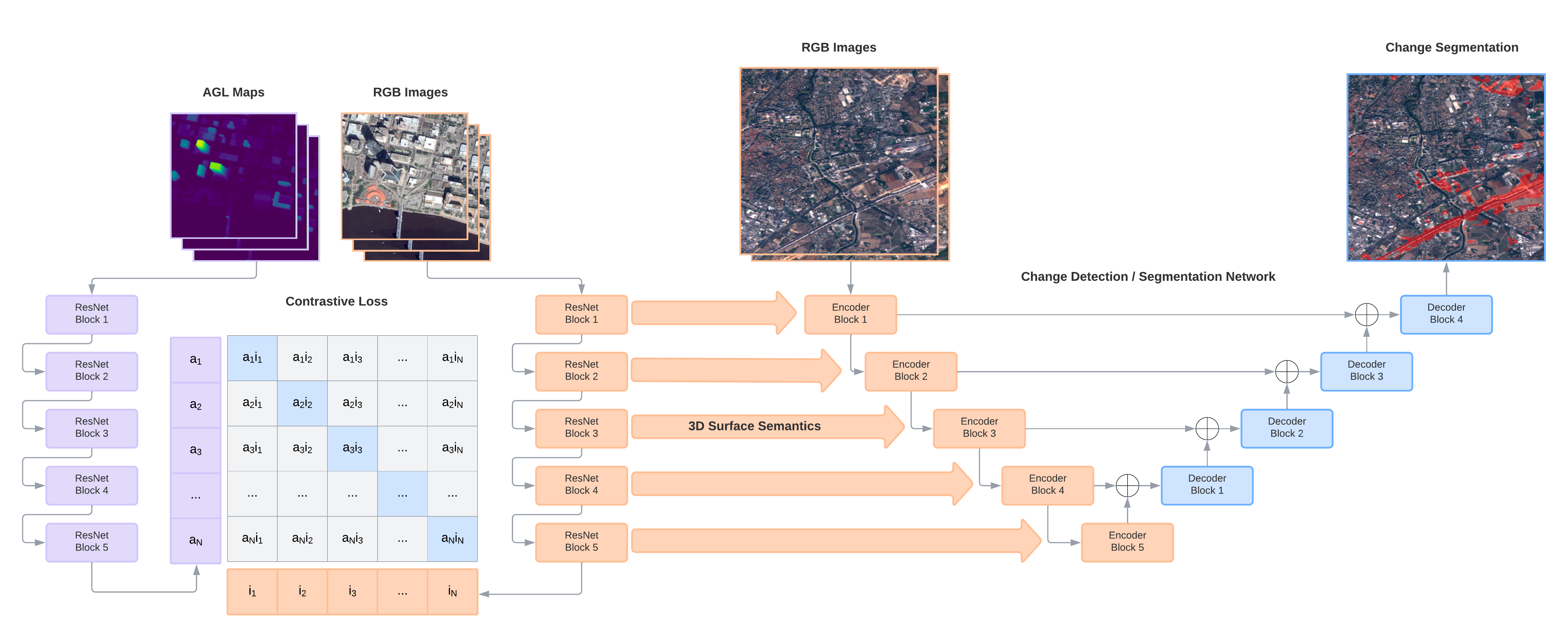}}
\caption{An overview of the CSIP architecture we propose. Using optical RGB and AGL map pairs, the multimodal self-supervised pretext encoders contrastively learns to extract features relevant to 3D surfaces, after which a downstream model provides change detection segmentations based on the learned 3D surface semantics.}
\label{fig:csip2}
\end{figure*}

Change detection is an important problem in remote sensing. The development of large scale labeled change detection datasets has allowed for rapid humanitarian assistance and disaster response (HADR) applications for natural disaster relief. Improvements in performance of computer vision models for change detection is critical for planning and accurate detection of areas requiring assistance from the HADR organizations and communities.

The combination of self-supervised learning methods for change detection has been explored recently using mainly using pre and post imagery as multiple views of the same location. Leenstra et al. \cite{leenstra2021self} explored using a large dataset of Sentinel-2 imagery sampled over the same geographic location at different times to train a model to predict whether a patch spatially overlaps another patch as a pretext task. Chen et al. \cite{chen2021self} used pre and post imagery to train using a contrastive framework. However, to our knowledge, no other methods have contrastively trained using overlapping remotely sensed imagery and 3D surface maps as a pretext task.

Our contributions can be described as following:

\begin{itemize}
\item We propose a contrastive multimodal framework for pretraining deep neural networks to learn representations of remotely sensed imagery which extract features relevant to 3D surface information.
\item We perform thorough investigation through extensive experiments on downstream tasks to validate our hypothesis that our pretraining framework does in fact improve performance on tasks which require surface or height extraction to perform image segmentation.
\end{itemize}

\section{Methods}
\label{sec:methods}
\begin{figure*}[ht!]
  \centering
  \centerline{\includegraphics[width=\textwidth]{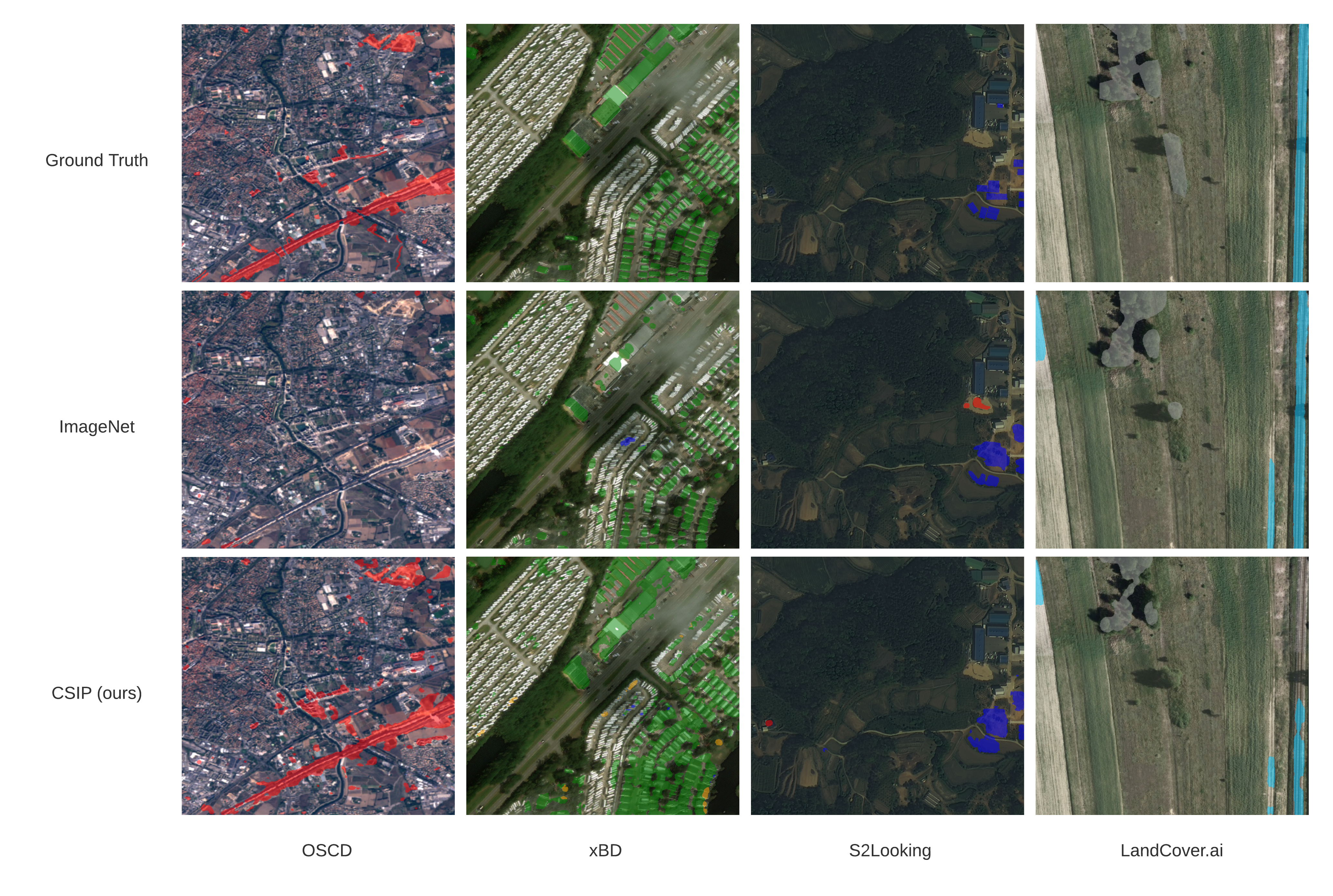}}
\caption{Visual comparisons of ImageNet and CSIP weight initializations to ground truth on randomly sampled test set predictions.}
\label{fig:results}
\end{figure*}

Inspired by the Contrastive Language-Image Pretraining (CLIP) \cite{clip} framework, we propose 
Contrastive Surface-Image Pretraining (CSIP) framework which utilizes dual encoders trained using a contrastive loss to encode imagery and above ground level (AGL) maps into similar latent space representations. Given an image $x_{i}$ and AGL map $x_{a}$ pair, an image encoder network $f_{i}$, and AGL encoder network $f_{a}$, we encode the image and AGL map into feature vectors $z_{i}$ and $z_{a}$, respectively. Each encoder network is composed of a backbone and a projection multi-layer perceptron (MLP) head. We then minimize the Normalized Temperature-scaled Cross Entropy (NT-Xent) loss \cite{sohn2016improved} with a learned temperature parameter $\tau$ which is a variant of the InfoNCE contrastive loss \cite{oord2018representation} and is described in Equation \ref{eq:loss}. By minimizing the NT-Xent loss we seek to maximize the cosine similarity between latent RGB and AGL vectors from the same pair relative to other latent vectors in a minibatch. Within a minibatch, we utilize other randomly sampled pairs as negative samples. An overview of the CSIP architecture is provided in Figure \ref{fig:csip}.

\begin{equation}
\label{eq:loss}
    \ell_{i,j} = -\log \frac{\exp(\mathrm{sim}(\bm z_i, \bm z_j)/\tau)}{\sum_{k=1}^{2N} \one{k \neq i}\exp(\mathrm{sim}(\bm z_i, \bm z_k)/\tau)}~,
\end{equation}

The end result of our framework is to take the pretrained image encoder network, $f_{i}$, and use as a backbone or encoder in segmentation or change detection networks. An overview of this is provided in Figure \ref{fig:csip2}.

\section{Experiments}
\label{sec:experiments}
To assess the value of the feature representations learned during pretraining, we compare against backbones pretrained on ImageNet \cite{deng2009imagenet} on several downstream tasks. In all experiments, we freeze the pretrained backbones and only fine-tune the model heads for each downstream task. In all downstream experiments we use a batch size of 16, 25 number of epochs, AdamW optimizer \cite{loshchilov2017decoupled} with a learning rate of $\alpha = 0.001$, patch sizes of 512x512, and a training/validation split of 0.2 for datasets which did not have a predefined validation split. For measuring downstream change detection and segmentation performance we evaluate using the mean Intersection-over-Union (mIoU), F1-score, and Average Accuracy metrics.

\subsection{Contrastive Pretraining}
During pretraining, we use the Overhead Geopose Challenge dataset \cite{christie2020learning} which is an extension to the Urban Semantic 3D (US3D) dataset \cite{us3d} used in the 2019 IEEE GRSS Data Fusion Contest (DFC19) \cite{dfc19}. The dataset is composed of pairs of 2048x2048 2D orthorectified RGB Maxar WorldView satellite imagery and 3D above ground level (AGL) maps. Furthermore, the dataset consists of oblique or significantly off-nadir imagery which adds additional complexity during pretraining. We pretrain separate ResNet-18 \cite{resnet} RGB and AGL encoders using the contrastive framework proposed in Section \ref{sec:methods} for 500 epochs. We then utilize the RGB encoder as a backbone in the following experiments.

\subsection{Change Detection}
We perform experiments for change detection on the following datasets:

\begin{description}
\item[OSCD] The Onera Satellite Change Detection (OSCD) \cite{daudt2018urban} dataset is a dataset for building change detection. The dataset contains 24 multispectral image pairs of various sizes taken by the Sentinel-2 satellites \cite{drusch2012sentinel} along with binary change masks representing building change between pre and post imagery.

\item[xBD] The xBD \cite{gupta2019creating} is a dataset for change detection and HADR applications utilized in the xView2 Challenge. The labels consist of building polygons categorized into 4 damage severities in imagery from before and after natural disasters.

\item[S2Looking] The S2Looking \cite{shen2021s2looking} dataset is a building change detection dataset consisting of 5,000 1024x1024 0.5-0.8 m resolution image pairs taken by the Gaofen, SuperView, and BeiJing-2 satellites. The dataset contains imagery with significantly off-nadir angles of rural locations along with labels for both newly built and demolished buildings.
\end{description}

\begin{table}[h]
\centering
\begin{tabular}{@{}ccccc@{}}
\toprule
\textbf{Dataset}        & \textbf{Weight Init.} & \multicolumn{1}{c}{\textbf{mIoU}} & \multicolumn{1}{c}{\textbf{F1}} & \multicolumn{1}{c}{\textbf{Acc.}} \\
\midrule
\multirow{2}{*}{OSCD}      & ImageNet    & 0.302          & 0.674          & 0.643          \\
                           & CSIP (ours) & \textbf{0.454} & \textbf{0.692} & \textbf{0.707} \\
\midrule
\multirow{2}{*}{xView2}    & ImageNet    & 0.315          & 0.526          & 0.480          \\
                           & CSIP (ours) & \textbf{0.367} & \textbf{0.529} & \textbf{0.533} \\
\midrule
\multirow{2}{*}{S2Looking} & ImageNet    & 0.415          & \textbf{0.602} & 0.609          \\
                           & CSIP (ours) & \textbf{0.490} & 0.568          & \textbf{0.657} \\
\bottomrule
\end{tabular}
\caption{
Experimental results on three benchmark change detection datasets comparing our CSIP method with ImageNet pretrained weights in the encoder of FCSiamDiff. Best results are marked in bold.}
\label{tab:results}
\end{table}

During training we fine-tune the Fully-Convolutional Siamese Difference (FC-Siam-Diff) \cite{daudt2018fully} architecture which is based on the U-Net architecture with the exception of taking the difference between skip connections after passing both pre and post imagery through the U-Net encoder. To compare the learned semantic surface features, we freeze the encoder of the network and only fine-tune the decoder on each dataset.

\subsection{Semantic Segmentation}
To compare performance on tasks not focused solely on building change or mapping, we conduct experiments on the LandCover.ai land cover semantic segmentation dataset \cite{Boguszewski_2021_CVPR}. The dataset consists of 10,674 512x512 RGB aerial images with 0.25-0.55 m spatial resolution. The imagery is annotated with semantic masks of 4 land cover categories. For benchmarking purposes, we utilize the original splits provided with the dataset. During training, we fine-tune the U-Net \cite{ronneberger2015u} architecture with the pretrained ResNet backbone as the encoder. Similarly to the change detection experiments, we freeze the encoder of the network and only fine-tune the decoder on the dataset.

\begin{table}[h]
\centering
\begin{tabular}{@{}cccc@{}}
\toprule
\textbf{Weight Init.} & \multicolumn{1}{c}{\textbf{mIoU}} & \multicolumn{1}{c}{\textbf{F1}} & \multicolumn{1}{c}{\textbf{Acc.}} \\
\midrule
ImageNet & \textbf{0.833} & \textbf{0.865} & \textbf{0.860} \\
\midrule
CSIP (ours) & 0.766          & 0.843          & 0.809          \\ 
\bottomrule
\end{tabular}
\caption{
Experimental results on the LandCover.ai land cover semantic segmentation dataset comparing our CSIP method with ImageNet pretrained weights in the encoder of U-Net. Best results are marked in bold.}
\label{tab:segmentation}
\end{table}

\section{Discussion and Future Work}
\label{sec:discussion}
The experimental results in Table \ref{tab:results} provide evidence that pretraining using the CSIP framework does in fact improve downstream performance for building change detection datasets in comparison to ImageNet initialization. Additionally, the results on the LandCover.ai segmentation dataset display a limitation to the framework where surface and height information may not be as important and ImageNet feature extraction has a distinct advantage where texture and color is more important in the delineating between land cover categories. Visual comparisons on randomly sampled test set predictions is provided for all datasets in Figure \ref{fig:results}.

While our experiments only explore pretraining of ResNet backbones, we note that investigation of pretraining Vision Transformers (ViT) \cite{dosovitskiy2020image} would be a direct next step in this research. ViTs have shown improved performance in segmentation \cite{ranftl2021vision} and change detection \cite{chen2021remote} applications in comparison to fully-convolutional based architectures. Furthermore, pretraining of ViT backbones with self-supervision \cite{caron2021emerging} has recently shown promising results.

\section{Conclusion}
\label{sec:conclusion}
In this paper, we propose a contrastive framework for supervising optical RGB image feature extraction of 3D surface information using AGL maps. We then presented experimental results supporting the hypothesis that pretraining vision models with self-supervision from 3D surface maps would improve downstream performance for building change detection which has direct applications in HADR efforts.

\bibliographystyle{IEEEbib}
\bibliography{refs}

\end{document}